\documentclass[11pt]{article}

\usepackage[preprint]{acl}

\usepackage{times}
\usepackage{latexsym}

\usepackage[T1]{fontenc}

\usepackage[utf8]{inputenc}

\usepackage{microtype}

\usepackage{inconsolata}

\usepackage{graphicx}

\usepackage{multirow}
\usepackage{booktabs}
\usepackage{amsmath}
\usepackage{tcolorbox}

\title{

How Robust is OCR-Reasoning? Evaluating OCR-Reasoning Robustness of Vision-Language Models under Visual Perturbations

}

\author{Yuxing Cheng$^{1}$  \quad Yuan Wu$^{1}$\footnotemark[1] \quad  Yi Chang$^{1,2,3}$\footnotemark[1] \\
        $^{1}$School of Artificial Intelligence, Jilin University \\
        $^{2}$Engineering Research Center of Knowledge-Driven Human-Machine Intelligence, MOE, China \\
        $^{3}$International Center of Future Science, Jilin University\\
        chengyx0121@mails.jlu.edu.cn, yichang@jlu.edu.cn, yuanwu@jlu.edu.cn \\
}

\begin{document}
\maketitle

\footnotetext[1]{Corresponding authors}

\begin{abstract}
Vision-language models (VLMs) have achieved strong performance on OCR-based benchmarks and increasingly focused on text-rich understanding, but their robustness under controlled visual degradation remains insufficiently understood. This gap is critical for OCR reasoning, where visual corruption can induce OCR errors and structural distortions, thereby introducing uncertainty into the reasoning task. To systematically study this problem, we introduce OCR-Robust, a benchmark designed for evaluating OCR reasoning robustness under visual perturbations. It contains 812 samples across two complementary subsets: OCR1.0, covering documents, scene text, receipts, handwriting, and mathematical content, and OCR2.0, focusing on charts, geometry diagrams, and tables. To enable efficient yet informative evaluation, we conduct a pilot study over 18 candidate perturbations and select 5 representative types at 3 severity levels each based on their impact and cross-model discriminability. We evaluate robustness using clean accuracy, Relative Corruption Retention (RCR), Worst-Case Retention (WCR), and a composite Corruption Robustness Index (CRI), and benchmark 18 models spanning proprietary systems, open-source VLMs, and OCR+LLM pipelines. Our results show that higher clean accuracy does not necessarily imply stronger robustness, and that models can suffer pronounced degradation in the worst case on OCR tasks that are sensitive to structure, and charts and tables are substantially more fragile than document-like inputs under perturbation.
\end{abstract}

\section{Introduction}

Vision-Language Models (VLMs) have achieved remarkable performance on OCR and visual reasoning benchmarks. Models such as GPT-5.2~\citep{openai2025gpt5}, Qwen3-VL~\citep{qwen2025qwen3vl}, and Gemini-3-Flash~\citep{deepmind2025gemini3flash} demonstrate strong capabilities in document understanding, chart interpretation, and scene text recognition.
Specialized OCR systems such as HunyuanOCR~\citep{hunyuanocr}, PaddleOCR~\citep{cui2025paddleocr}, and DeepSeekOCR~\citep{wei2025deepseekocr} have achieved strong text extraction performance on clean images, with some reported results approaching or surpassing human performance on OCRBench~\citep{liu2024ocrbench} and OmniDocBench~\citep{ouyang2025omnidocbench}.
Yet on OCR-reasoning tasks that jointly demand text-rich perception and multi-step reasoning, current models still exhibit notable performance bottlenecks~\citep{ye2025logicocr,huang2025ocrreasoning}. 
Moreover, even these evaluations are conducted on clean and curated images. 
Practical deployment rarely presents such ideal conditions: documents are photographed under uneven lighting, receipts are crumpled and faded, and street signs are captured through rain or motion blur.

The problem extends beyond perceptual difficulty alone.
OCR reasoning robustness is fundamentally different from generic visual robustness: 
visual corruption does not merely blur semantic content, but can induce discrete symbolic errors that are subsequently amplified by downstream reasoning tasks. 

Existing benchmarks do not isolate this phenomenon. 
Standard OCR evaluations~\citep{fu2024ocrbench2,ouyang2025omnidocbench,huang2025ocrreasoning} focus on clean images. 
General robustness benchmarks such as ImageNet-C~\citep{hendrycks2019robustness}, 
VLM-RobustBench~\citep{saxena2026vlmrobustbench} assess object recognition under corruption but do not involve text extraction or reasoning. 
Even recent benchmarks for OCR robustness, such as CC-OCR~\citep{yang2025ccocr}, primarily evaluate robustness on general OCR tasks rather than OCR reasoning under visual perturbations.
To systematically study this problem, we introduce \textbf{OCR-Robust}, a benchmark designed for evaluating OCR reasoning robustness under visual perturbations. 
Rather than serving as a large-scale data resource, OCR-Robust extends the corruption robustness evaluation paradigm to OCR reasoning, providing a controlled diagnostic setting for studying how visual perturbations affect text-rich perception and downstream reasoning.
We do not aim to simulate real-world image degradation, and our compound perturbation setting should be viewed as an initial attempt to probe interactions among controlled visual corruptions.
Code and supplementary materials are available at \url{https://github.com/pasterinjlu/OCR-Reasoning-Robust}. 
Our benchmark makes three contributions:

\begin{itemize}
    \item \textbf{A diverse, two-part dataset.} We construct 812 samples spanning two complementary subsets: OCR1.0 (482 samples) covers documents, scene text, receipts, handwriting, screenshots, and mathematical content; OCR2.0 (330 samples) targets charts, geometry diagrams, and tables. Together, they cover the full spectrum from natural text perception to structured visual reasoning.

    \item \textbf{A principled perturbation selection pipeline.} Rather than applying perturbations ad hoc, we conduct a pilot study evaluating 18 candidate perturbation types across 6 models, using discriminability metrics to select the 5 perturbations that may more effectively differentiate model robustness: glass blur, motion blur, elastic deformation, color shift, and snow. Each is applied at 3 severity levels, yielding 15 evaluation conditions.

    \item \textbf{A robustness metric suite.} We propose clean accuracy and three complementary metrics: Relative Corruption Retention (RCR), Worst-Case Retention (WCR), and a composite Corruption Robustness Index (CRI) that balances baseline capability, average retention, and worst-case risk through geometric mean aggregation.
\end{itemize}

Through extensive evaluation of 18 models spanning proprietary systems (GPT-5.2~\citep{openai2025gpt5}, Gemini-3-Flash~\citep{deepmind2025gemini3flash}), open-source VLMs (Qwen3-VL~\citep{qwen2025qwen3vl}, InternVL3.5~\citep{zhu2025internvl3}), and OCR+LLM pipelines (PaddleOCR-VL-1.5~\citep{cui2026paddleocrvl15} + GPT-5.1), 
we observe five notable findings: 
(1) closed-source models tend to lead in overall robustness; 
(2) higher capability does not guarantee stronger robustness, as thinking mode models achieve the highest clean accuracy but still show large drops on structured content, with WCR as low as 0.579;
(3) structured visual content (charts, tables) appears substantially more fragile than documents and scene text, with average worst-case retention declining from 0.826 on OCR1.0 to 0.676 on OCR2.0; 
(4) OCR+LLM pipelines are sensitive to OCR extraction quality;
(5) prompt-induced chain-of-thought (CoT) improves clean accuracy but does not reliably prevent worst case failures, suggesting that symbolic brittleness may be most pronounced where reasoning demands are highest.

\section{Related Work}

\subsection{OCR and Text-Rich Benchmarks}

VLMs are expected not only to recognize text but also to localize, parse layouts, and reason over textual evidence. 
This has driven the development of comprehensive OCR benchmarks. OCRBench~\cite{liu2024ocrbench} and its successor OCRBench v2~\cite{fu2024ocrbench2} cover tasks from basic recognition to complex visual text reasoning. 
CC-OCR~\cite{yang2025ccocr} targets multilingual reading and document parsing, while OmniDocBench~\cite{ouyang2025omnidocbench} focuses on detailed PDF understanding. 
For long documents, MMLongBench~\cite{wang2025mmlongbench} and M-LongDoc~\cite{chia2025mlongdoc} require cross-page evidence integration.
More recently, benchmarks have shifted toward OCR reasoning: LogicOCR~\cite{ye2025logicocr} tests logical inference over text-rich images, 
while OCR-Reasoning~\cite{huang2025ocrreasoning} categorizes evaluations by fine-grained reasoning abilities. 

\subsection{Robustness Evaluation for VLMs}

Recent work has increasingly examined model stability under naturalistic corruptions.
\citet{ishmam2025visual} stress-tests VQA models with 213K augmented images and proposes aggregatable robustness metrics. 
MLLM-IC~\cite{qiu2025mllmcorruption} evaluates models across 40 corruption types and 34 capabilities for fine-grained diagnosis. 
Besides comprehensive variation types~\cite{saxena2026vlmrobustbench,fan-etal-2025-unveiling}, other benchmarks address multi-modal shifts~\cite{huang2025frames}, self-consistency under corruption~\cite{zhang2024mmcbench}, resolution robustness~\cite{li2025resbench}, and coordinated cross-modal perturbations~\cite{babu2025coordinated}.


\section{Base Benchmark Construction}

We construct two complementary datasets combining multiple sources to ensure comprehensive coverage of OCR reasoning scenarios. 
Our approach balances efficiency by leveraging existing high quality datasets, coverage through targeted annotation, and quality via human review of VLM generated questions. 
These datasets serve as the base set for our robustness evaluation. More construction details provided in Appendix~\ref{sec:appendix_data_construction}.

\subsection{Data Sources and Composition}

Our benchmark comprises 812 samples across two complementary datasets, as summarized in Table~\ref{tab:data_sources}.
OCR1.0 (482 samples) spans nine data types covering documents, scene text, receipts, handwriting, and mathematical content, focusing on multi-hop reasoning and cross-reference integration.
OCR2.0 (330 samples) draws from ChartVQA~\citep{masry2022chartqa}, GNS-260K~\citep{ning2025gns}, and TableVQA-Bench~\citep{kim2024tablevqa}, targeting chart, geometry, and table comprehension 
that requires analysis of structured visual content, such as trend analysis, value aggregation, and spatial reasoning.
Together, the two subsets cover the main spectrum from document-level text extraction to structured visual reasoning. 

\begin{table}[htbp]
\centering
\small
\begin{tabular}{@{}ll@{\hspace{6pt}}l@{\hspace{6pt}}r@{}}
\toprule
Dataset & Data Type & Source & \# \\
\midrule
\multirow{9}{*}{OCR 1.0}
 & Eng.\ Document & DocVQA & 50 \\
 & Scene Text & TextVQA & 52 \\
 & Infographic & InfographicVQA & 40 \\
 & Chn.\ Advertisement & MTWI & 63 \\
 & Receipt & SROIE & 75 \\
 & Screenshot & Screen2Words & 30 \\
 & Handwritten & OCR-Reasoning & 30 \\
 & Mathematical & GSM8K/TheoremQA & 100 \\
 & Chn.\ Document & M6Doc & 42 \\
\midrule
\multirow{3}{*}{OCR 2.0}
 & Chart & ChartVQA & 150 \\
 & Geometry & GNS-260K & 80 \\
 & Table & TableVQA-Bench & 100 \\
\midrule
\multicolumn{3}{@{}l}{Total} & 812 \\
\bottomrule
\end{tabular}

\caption{Distribution of data types in the benchmark.}
\label{tab:data_sources}
\end{table}

\subsection{Construction Methods}

We employ three complementary approaches tailored to different data characteristics:

\paragraph{Direct Selection}
For OCR2.0 and portions of OCR1.0 (handwriting from OCR-Reasoning~\citep{huang2025ocrreasoning} and InfographicVQA~\citep{mathew2021infographicvqa} subsets), we select samples from existing high-quality datasets. OCR2.0 employs visual diversity sampling to ensure broad coverage of chart types. 
We extract 2048-dimensional ResNet50~\citep{he2015resnet} features from candidate images, apply agglomerative clustering~\citep{2011clustering} with Ward linkage to form 150 clusters, and select the sample closest to each cluster centroid. 
This automated approach efficiently identifies diverse samples while preserving original questions and answers.

\paragraph{VLM-Assisted Annotation}
For OCR1.0, the DocVQA~\citep{mathew2020docvqa}, TextVQA~\citep{singh2019textvqa}, MTWI~\citep{2018mtwi}, SROIE~\citep{Huang_2019_sroie}, Screen2Words~\citep{wang2021screen2wordsautomaticmobileui}, and M6Doc~\citep{Cheng_2023_m6doc} subsets undergo VLM-assisted annotation to generate reasoning questions. We use three strong VLMs: GPT-5.2~\citep{openai2025gpt5}, Gemini-3-Pro~\citep{deepmind2025gemini3pro}, and Claude-Sonnet-4.5~\citep{anthropic2025claudesonnet45}. 
For each image, one randomly selected model generates a reasoning question and provides the corresponding answer. 
Using multiple models also helps mitigate potential data contamination issues. 
All VLM-Assisted generated question-answer pairs are verified by human annotators to ensure quality and correctness.

\paragraph{Synthetic Generation}
The math subset in OCR1.0 is synthetically constructed from GSM8K~\citep{cobbe2021gsm8k} and TheoremQA~\citep{chen-etal-2023-theoremqa}. 
Following the LogicOCR~\citep{ye2025logicocr} pipeline,
we first convert multiple choice questions into free-form answer format to better evaluate reasoning ability. 
Then we use GPT-4o to generate a visual context description for each question, and leverage GPT-4o-image to produce the corresponding image with the context and question rendered into it. 
This pipeline naturally provides ground-truth answers, and we further use PaddleOCR~\citep{cui2025paddleocr} to compute character recall for filtering.

\subsection{Quality Control}

We enforce multiple quality control criteria throughout dataset construction. 
First, we manually verify both answer verifiability and OCR necessity, ensuring that each selected sample has an objectively supported answer grounded in the image content and cannot be solved without text extraction. 
Second, we perform near-duplicate removal based on perceptual image hashing.
Finally, we perform a final manual review of annotated samples to ensure correctness detailed in Appendix~\ref{sec:appendix_human}.
Only samples that pass all quality criteria are retained in the final benchmark.

\section{Experiments}

\subsection{Experimental Setup}

\paragraph{Evaluated Models}

We evaluate three categories of systems on OCR-Robust: closed-source VLMs, open-source VLMs, and OCR+LLM pipelines.

\textbf{(a) Open-source MLLMs.}
We assess a broad range of open-source models spanning diverse scales and architectures: Qwen2.5-VL-7B~\citep{qwen25vl}, Qwen3-VL-4B/8B/32B-Instruct~\citep{qwen2025qwen3vl}, Qwen3-VL-4B/8B-Thinking~\citep{qwen2025qwen3vl}, InternVL3.5-4B/8B/14B/38B~\citep{zhu2025internvl3}, Llama-4-Scout-17B-16E-Instruct~\citep{meta2025llama4}, Kimi-VL-A3B-Thinking~\citep{team2025kimi}, VL-Rethinker-7B~\citep{wang2025vlrethinker}, and MM-Eureka-Qwen-7B~\citep{meng2025mmeureka}. Among these, the Thinking variants employ extended chain-of-thought reasoning during inference.
\textbf{(b) Closed-source MLLMs.}
We also evaluate leading closed-source models, including GPT-5.2~\citep{openai2025gpt5}, GPT-5.1~\citep{openai2025gpt5}, and Gemini-3-Flash~\citep{deepmind2025gemini3flash} (with and without thinking mode).
\textbf{(c) OCR + LLM Pipelines.}
Finally, to examine whether decoupling perception from reasoning offers robustness advantages, we evaluate a two-stage pipeline where PaddleOCR-VL-1.5 ~\citep{cui2026paddleocrvl15} first extracts text from images, which is then fed to a LLM for text-only reasoning. We test GPT-5.1~\citep{openai2025gpt5} and Gemini-3-Flash (thinking mode)~\citep{deepmind2025gemini3flash} as downstream reasoning models.

\paragraph{Implementation Details}
We adopt a zero-shot evaluation protocol without fine-tuning or few-shot prompting. 
For VLMs, we present both the image and the textual question, appended with a format hint from the dataset annotation (e.g., ``The composition of the final answer should be: \$ + Integer'') to ensure a consistent output format. 
The prompt ends with ``Directly output the answer only, without any explanation.'' 
For OCR + LLM pipelines, 
we replace the image input with text extracted by PaddleOCR-VL-1.5 while keeping the same query structure.
Scoring uses normalized exact match, with GPT-4o as a semantic fallback judge, as detailed in Appendix~\ref{sec:eval_protocol}.

\paragraph{Metrics}
\label{sec:metrics}

We evaluate robustness from four complementary perspectives. Let $A^{\text{clean}}_{m,d}$ denote the accuracy of model $m$ on dataset $d$ under clean images, and $A_{m,d,c,s}$ denote the accuracy under corruption family $c$ at severity level $s$. Let $\Omega_d = \{(c,s)\}$ be the set of all observed perturbation conditions on dataset $d$.

\noindent \textbf{Clean Accuracy} is the baseline accuracy on unperturbed images, reported separately to distinguish inherent capability from robustness.

\noindent\textbf{Relative Corruption Retention (RCR)} measures the average ratio of corrupted accuracy to clean accuracy:

\begin{equation}
    \mathrm{RCR}(m,d) =
    \frac{1}{|\Omega_d|}
    \sum_{(c,s) \in \Omega_d}
    \min\left(
        \frac{A_{m,d,c,s}}{A^{\text{clean}}_{m,d}},
        1
    \right)
\end{equation}
A value of 1 indicates perfect retention. 
Unlike mean corruption accuracy (mCA), RCR normalizes by baseline performance, 
enabling fair comparison between models of different capability levels.

\noindent\textbf{Worst-Case Retention (WCR)} captures the most severe failure mode by normalizing the minimum corrupted accuracy by the clean baseline:
\begin{equation}
    \mathrm{WCR}(m,d) = \frac{\min_{(c,s) \in \Omega_d} A_{m,d,c,s}}{A^{\text{clean}}_{m,d}}
\end{equation}
A model with high RCR but low WCR has a hidden vulnerability in specific corruption types.

\noindent\textbf{Corruption Robustness Index (CRI)} balances clean capability, average retention, and worst-case risk via the geometric mean of three factors:
\begin{equation}
    \mathrm{CRI}(m,d) = \sqrt[3]{\, \tfrac{A^{\text{clean}}_{m,d}}{100} \cdot \mathrm{RCR}(m,d) \cdot \mathrm{WCR}(m,d) \,}
\end{equation}
\noindent The geometric mean (cube root of the product) penalizes imbalance: a model that excels in one dimension but fails in another receives a lower CRI than one that performs consistently across all three, preventing high baseline accuracy from masking poor robustness.
$A^{\text{clean}}_{m,d}$ is divided by 100 for normalization.

\subsection{Perturbation Selection}

Not all visual perturbations are equally suitable for OCR reasoning robustness evaluation. 
A useful perturbation should jointly satisfy several criteria: it should meaningfully affect performance, distinguish models with different robustness levels, induce monotonic degradation as severity increases, complement rather than duplicate other perturbations. 
We therefore conduct a pilot study over 18 candidate perturbations across multiple VLMs to identify an informative subset.
Perturbation selection is part of the evaluation design, not a separate post-hoc experiment.

\paragraph{Experimental Setup}
We evaluated 6 VLMs covering closed-source (GPT-5.2), open-source (InternVL3.5-8B, Qwen2.5-VL-7B, Qwen3-VL-4B/8B-Instruct), and thinking-mode (Qwen3-VL-8B-Thinking) paradigms across \textbf{18 candidate perturbation types} in 5 categories: Noise (Gaussian, shot, impulse), Blur (defocus, glass, motion, zoom), Weather (fog, frost, snow), Digital (brightness, contrast, elastic, pixelate), and Color \& Geometric (color shift, glare, moir\'{e}, slight rotation). Each perturbation was applied at 4 severity levels controlled by LPIPS~\citep{zhang2018lpips} distance (0.05, 0.10, 0.15, 0.20), ensuring perceptually uniform degradation. We used a stratified sample of 100 questions from the OCR-Reasoning dataset~\citep{huang2025ocrreasoning}, covering diverse visual contexts.

\paragraph{Selection Metrics}

To reduce bias, our selection uses three complementary criteria, MRD, SEP, and MON.

Let $r_{m,c,s} = A_{m,c,s} / A^{\text{clean}}_m$ denote the retention ratio of model $m$ under perturbation $c$ at severity $s$, 
with $M$ models and $S_c$ severity levels per perturbation.

\noindent\textbf{Mean Relative Drop (MRD)} measures the average performance degradation a perturbation induces across all models and severity levels:

{\small
\begin{equation}
\text{MRD}(c) = \frac{1}{M \cdot |S_c|} \sum_{m,s} (1 - r_{m,c,s})
\end{equation}
}

\noindent Higher MRD indicates a stronger perturbation. However, high impact alone is insufficient: a perturbation is not informative if it degrades all models in nearly the same way.

\smallskip
\noindent \textbf{Separability (SEP)} quantifies how effectively a perturbation differentiates models with distinct robustness profiles, measured as the average pairwise retention divergence:

{\small
\begin{equation}
\text{SEP}(c) = \frac{1}{|S_c|} \sum_{s \in S_c} \frac{2}{M(M{-}1)} \sum_{i < j} |r_{i,c,s} - r_{j,c,s}|
\end{equation}
}

\noindent High SEP ensures the perturbation reveals meaningful differences between models rather than affecting all models equally.

\smallskip
\noindent\textbf{Monotonicity (MON)} measures whether performance degrades consistently as severity increases, rather than fluctuating erratically:

{\small
\begin{equation}
\text{MON}(c) = 1 - \frac{1}{M(|S_c|{-}1)} \sum_{m} \sum_{s} \mathbf{1}[r_{m,c,s+1} > r_{m,c,s}]
\end{equation}
}

\noindent A perturbation with high MON produces a reliable severity gradient, enabling fine-grained robustness profiling.

\begin{figure*}[t]
\centering
\includegraphics[width=0.75\textwidth]{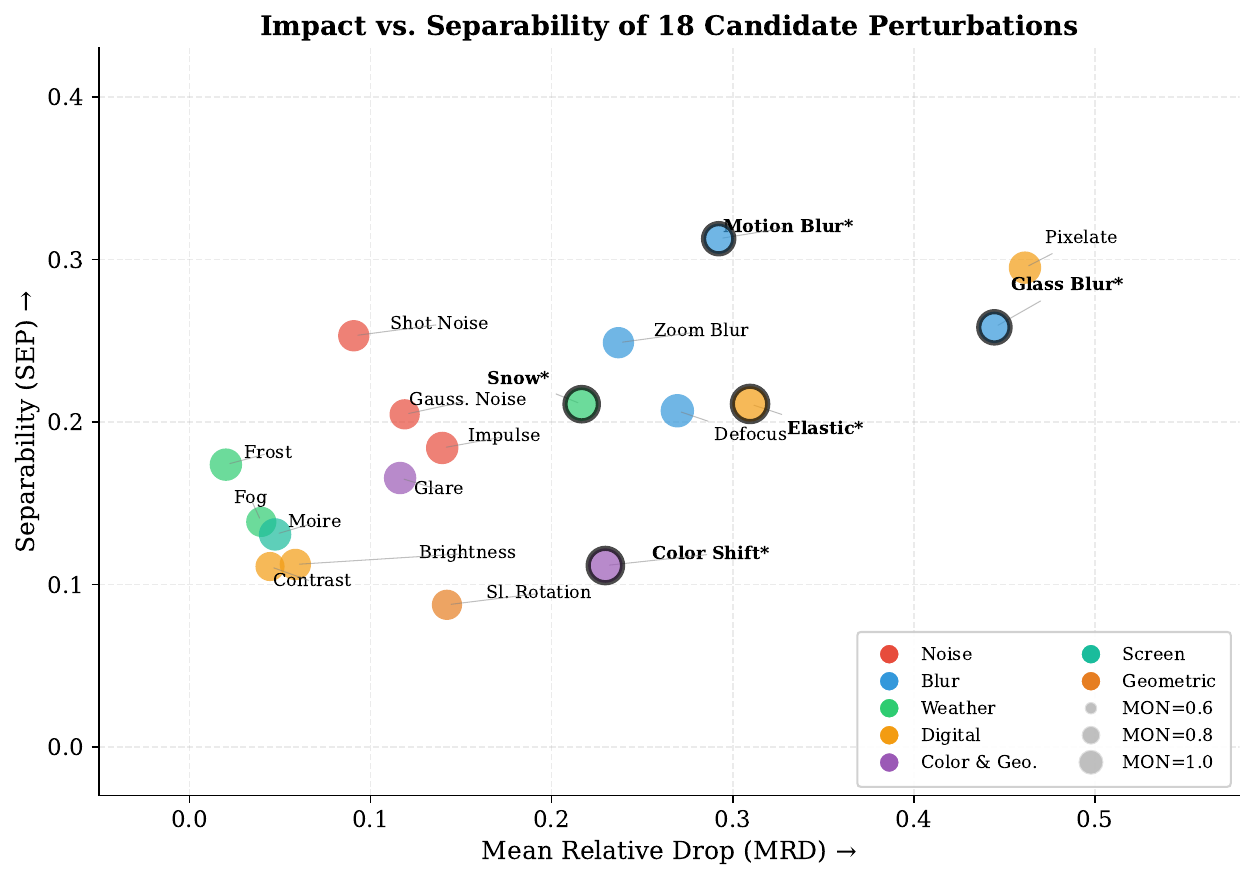}
\caption{Impact (MRD) vs.\ Separability (SEP) for 18 candidate perturbations. Point size encodes Monotonicity (MON), color indicates corruption category. Black circles mark the 5 selected perturbations.}
\label{fig:perturbation_scatter}
\end{figure*}

\paragraph{Results}
Table~\ref{tab:perturbation_ranking} and Figure~\ref{fig:perturbation_scatter} present the three selection metrics for all 18 candidates. 
As Figure~\ref{fig:perturbation_scatter} illustrates, several alternative perturbations are reasonably competitive. 
Rather than selecting perturbations based on a single ranking,
we first filter out candidates that catastrophically destroy text information rather than degrading it gradually. 
For example, pixelate is excluded despite its highest MRD of 0.461, as it renders characters unrecognizable at moderate severity and makes the task effectively unsolvable. 
We then apply greedy selection to the remaining candidates, selecting five perturbations in order:
\textbf{glass blur} (MRD\,=\,.445, SEP\,=\,.258, MON\,=\,.667) as the highest-impact non-destructive perturbation, \textbf{motion blur} (.292, .313, .611) for its strong separability, \textbf{elastic deformation} (.310, .211, .889) for its high monotonicity and low redundancy with blur types, \textbf{color shift} (.230, .112, .833) to cover color-based degradation, and \textbf{snow} (.217, .211, .778) to represent weather-induced corruption. 
The final subset spans four distinct corruption families while maintaining reasonable scores across all three axes. 
We note that alternative subsets with comparable overall quality are possible; 
our selection prioritizes category diversity and metric balance rather than claiming unique optimality. 
Bootstrap stability analysis (1,000 resamples, Appendix~\ref{sec:bootstrap_stability}) shows that these five perturbations are among the most frequently chosen, with selection rates ranging from 44.2\% to 98.4\%, though the moderate rate for color shift (44.2\%) indicates that substitution with other color-category perturbations could yield a comparably effective subset.

\begin{table}[t]
\centering
\small
\begin{tabular}{lcccc}
\hline
\textbf{Perturbation} & \textbf{MRD}$\uparrow$ & \textbf{SEP}$\uparrow$ & \textbf{MON}$\uparrow$ & \textbf{Cat.} \\
\hline
Pixelate & .461 & .295 & .722 & Digital \\
\textbf{Glass Blur}* & .445 & .258 & .667 & Blur \\
\textbf{Elastic}* & .310 & .211 & .889 & Digital \\
\textbf{Motion Blur}* & .292 & .313 & .611 & Blur \\
Defocus Blur & .269 & .207 & .778 & Blur \\
Zoom Blur & .237 & .249 & .667 & Blur \\
\textbf{Color Shift}* & .230 & .112 & .833 & Color \\
\textbf{Snow}* & .217 & .211 & .778 & Weather \\
Slight Rotation & .142 & .087 & .611 & Geometric \\
Impulse Noise & .140 & .184 & .722 & Noise \\
Gaussian Noise & .119 & .205 & .611 & Noise \\
Glare & .116 & .165 & .722 & Color \\
Shot Noise & .091 & .253 & .667 & Noise \\
Brightness & .058 & .112 & .667 & Digital \\
Moir\'{e} & .047 & .131 & .722 & Screen \\
Contrast & .044 & .111 & .556 & Digital \\
Fog & .040 & .138 & .611 & Weather \\
Frost & .020 & .174 & .722 & Weather \\
\hline
\end{tabular}
\caption{Perturbation selection metrics across 6 pilot models. * denotes the final 5 perturbations.}
\label{tab:perturbation_ranking}
\end{table}

\paragraph{Final Perturbation Configuration}
Each selected perturbation is configured with 3 severity levels: \textbf{Glass blur} (perturbation radius: 2, 2.5, 3 pixels), \textbf{Color shift} (channel offset: 3, 4, 5 pixels), \textbf{Elastic transform} (deformation strength $\alpha$: 10, 15, 20), \textbf{Motion blur} (kernel size: 5, 6, 7 pixels), and \textbf{Snow} (layer intensity: 0.1, 0.2, 0.3). These parameters are calibrated to produce perceptually meaningful degradation while maintaining text legibility at lower severity levels, yielding 5 dimensions $\times$ 3 levels = 15 perturbation conditions.
The pilot study involves 100 questions and 6 models, so the selected perturbation set should be viewed as an empirically supported protocol for this study, not as a universally optimal set. 

\begin{table*}[t]
\centering
\small
\begin{tabular}{@{}lcccccccc@{}}
\toprule
& \multicolumn{4}{c}{\textbf{OCR1.0}} & \multicolumn{4}{c}{\textbf{OCR2.0}} \\
\cmidrule(lr){2-5} \cmidrule(lr){6-9}
\textbf{Model} & Clean & RCR$\uparrow$ & WCR$\uparrow$ & CRI$\uparrow$ & Clean & RCR$\uparrow$ & WCR$\uparrow$ & CRI$\uparrow$ \\
\midrule
\multicolumn{9}{c}{\textit{Closed-source Models}} \\
\midrule
Gemini-3-Flash-Thinking & 79.05 & \textbf{.998} & \textbf{.932} & \textbf{.902} & 69.70 & .977 & \textbf{.870} & .840 \\
Gemini-3-Flash          & \textbf{80.71} & .986 & .920 & .901 & 71.52 & \textbf{.980} & .860 & \textbf{.845} \\
GPT-5.2                 & 74.69 & .911 & .799 & .816 & \textbf{71.82} & .876 & .717 & .767 \\
GPT-5.1                 & 60.58 & .986 & .882 & .808 & 49.70 & .931 & .762 & .706 \\
\midrule
\multicolumn{9}{c}{\textit{Open-source Models}} \\
\midrule
Qwen3-VL-8B-Thinking    & 79.67 & .961 & .875 & .875 & \textbf{78.18} & .892 & .624 & .758 \\
Qwen3-VL-4B-Thinking    & 78.63 & .948 & .859 & .862 & 76.97 & .879 & .579 & .732 \\
Qwen3-VL-32B-Instruct   & 72.82 & .954 & .826 & .831 & 57.27 & .928 & .640 & .698 \\
Qwen3-VL-8B-Instruct    & 54.15 & .903 & .770 & .722 & 49.39 & .921 & .632 & .660 \\
Qwen3-VL-4B-Instruct    & 49.79 & .927 & .788 & .714 & 49.70 & .841 & .555 & .615 \\
\cmidrule(l){1-9}
InternVL3.5-38B         & 62.86 & .933 & .794 & .775 & 61.52 & .941 & .650 & .722 \\
InternVL3.5-14B         & 55.81 & .939 & .838 & .760 & 55.45 & .912 & .738 & .720 \\
InternVL3.5-8B          & 57.26 & .941 & .835 & .766 & 56.67 & .892 & .631 & .683 \\
InternVL3.5-4B          & 52.49 & .935 & .796 & .731 & 51.52 & .905 & .647 & .671 \\
\cmidrule(l){1-9}
Llama-4-Scout-17B       & 67.84 & .939 & .867 & .821 & 59.70 & .866 & .711 & .716 \\
MM-Eureka-Qwen-7B       & 51.25 & .918 & .789 & .719 & 47.58 & .888 & .637 & .646 \\
VL-Rethinker-7B         & 52.49 & .914 & .776 & .719 & 48.18 & .928 & .610 & .649 \\
\midrule
\multicolumn{9}{c}{\textit{OCR + LLM Pipelines}} \\
\midrule
PaddleOCR-VL-1.5 + GPT-5.1  & 52.90 & .902 & .753 & .711 & 26.36 & .982 & .805 & .593 \\
PaddleOCR-VL-1.5 + Gemini-3-Flash-T & 65.15 & .938 & .796 & .787 & 48.79 & .910 & .665 & .666 \\
\bottomrule
\end{tabular}
\caption{Robustness evaluation on OCR1.0 (482 samples) and OCR2.0 (330 samples). Models grouped by family, ranked by average CRI. Best values per dataset in \textbf{bold}.}
\label{tab:main_results}
\end{table*}


\subsection{Results and Analysis}

Table~\ref{tab:main_results} presents the robustness evaluation results on both OCR1.0 and OCR2.0. Models are grouped by family and ranked by average CRI.
Meanwhile, we conducted a potential data contamination analysis in Appendix~\ref{sec:appendix_contaminaiton} and few-shot ablation study in Appendix~\ref{sec:appendix_few_shot}.

\begin{figure*}[t]
\centering
\includegraphics[width=\textwidth]{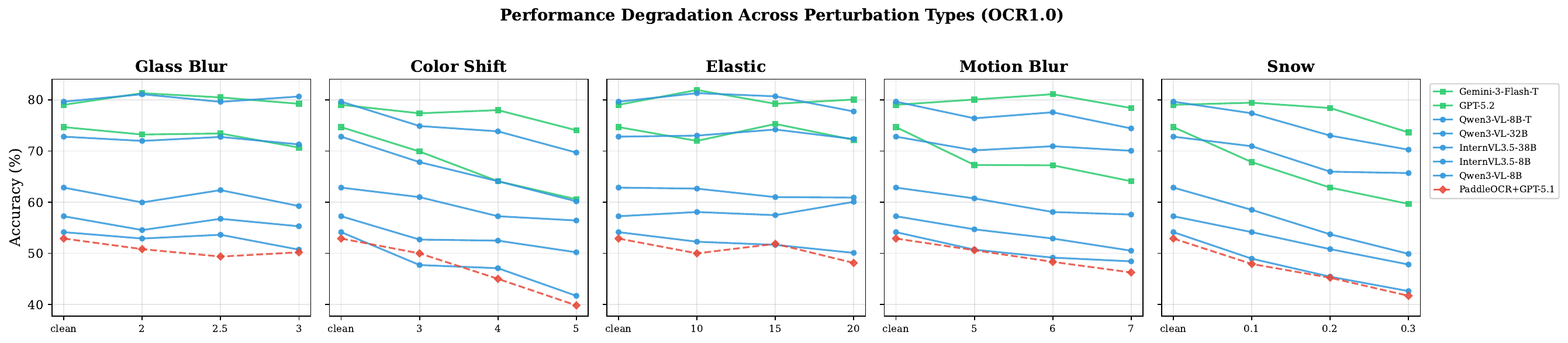}
\caption{Performance degradation across five perturbation types on OCR1.0. Each subplot shows accuracy from clean (leftmost) to highest severity (rightmost). Closed-source models (green squares) maintain flatter curves, while pipelines (red diamonds) exhibit steeper drops.}
\label{fig:collapse_ocr10}
\end{figure*}

\begin{figure*}[t]
\centering
\includegraphics[width=\textwidth]{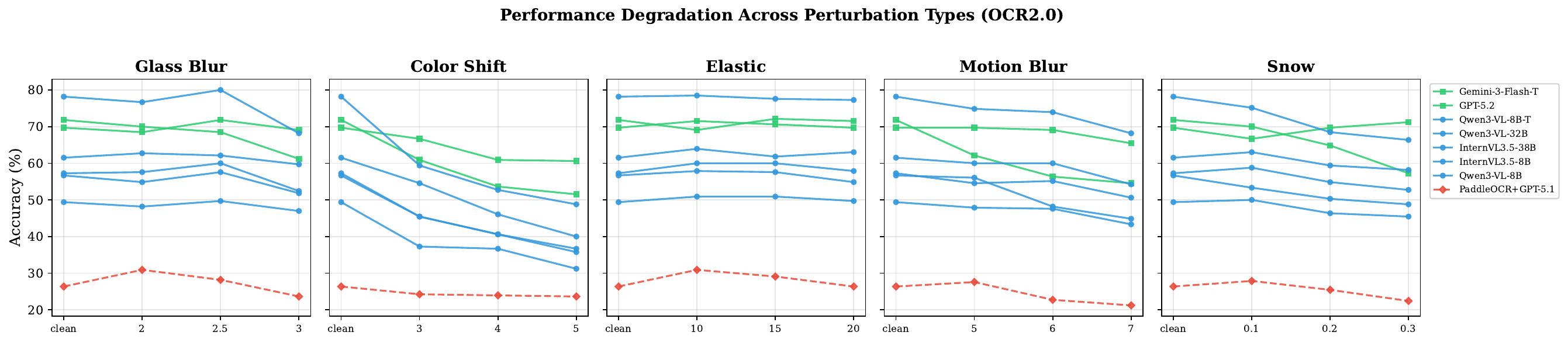}
\caption{Performance degradation across five perturbation types on OCR2.0. Compared to OCR1.0, all models show sharper collapses, particularly under color shift and snow perturbations.}
\label{fig:collapse_ocr20}
\end{figure*}

\paragraph{Finding 1: Among the evaluated models, closed-source models lead in overall robustness.}
As shown in Figure~\ref{fig:clean_vs_wcr}, the top-performing closed-source models occupy the upper-right quadrant, achieving CRI above 0.90 on OCR1.0 with RCR $\geq$ 0.986 and WCR $\geq$ 0.920, and lead on OCR2.0 with CRI above 0.84. These models maintain near-perfect average retention while exhibiting the strongest worst-case resilience, suggesting more robust visual understanding rather than merely higher baseline accuracy. Among open-source models, Qwen3-VL-8B-Thinking achieves the closest CRI (0.875 on OCR1.0), but its worst-case retention on OCR2.0 (0.624) reveals a significant gap compared to closed-source systems.

\paragraph{Finding 2: Higher capability does not guarantee higher robustness.}
This pattern manifests in two dimensions (see also the collapse curves in Figures~\ref{fig:collapse_ocr10} and~\ref{fig:collapse_ocr20}). First, \textit{native thinking models improve clean accuracy but not robustness}: Qwen3-VL-8B-Thinking, which employs internalized chain-of-thought learned during training, achieves the highest clean accuracy on OCR2.0 (78.18\%) yet its WCR drops to 0.624, while Gemini-3-Flash at 71.52\% clean maintains WCR of 0.860. The same Thinking models show much milder drops on OCR1.0 (WCR $\geq$ 0.859), suggesting that internalized reasoning is more vulnerable on structured visual content. Second, \textit{larger scale does not monotonically improve robustness}: within the InternVL3.5 family, the 14B model (WCR = 0.838 on OCR1.0, 0.738 on OCR2.0) consistently outperforms the 38B variant (WCR = 0.794 and 0.650) in worst-case retention despite lower clean accuracy. 
This suggests that scaling alone does not guarantee better corruption robustness.

\paragraph{Finding 3: Structured visual content is substantially more fragile in OCR-Robust.}
Across all models, WCR on OCR2.0 is consistently lower than on OCR1.0. The average WCR drops from 0.826 (OCR1.0) to 0.676 (OCR2.0), a relative decline of 18\%. 
Color shift emerges as one of the main drivers of this gap (Figure~\ref{fig:heatmap_drop}), with several models losing over 40\% of their clean accuracy at the highest severity on OCR2.0.
This suggests that charts, tables, and geometry diagrams are more sensitive to visual perturbations than natural scene text or documents, likely because structured visual content often depends on fine-grained visual cues.

\paragraph{
Finding 4: OCR+LLM pipelines are sensitive to OCR extraction quality.
}
The OCR + LLM pipelines exhibit a characteristic trade-off. PaddleOCR-VL-1.5 + GPT-5.1 achieves high RCR (0.902 on OCR1.0, 0.982 on OCR2.0) but the lowest CRI due to severely limited clean accuracy (52.90\% and 26.36\%). 
Replacing the reasoning backend with Gemini-3-Flash-Thinking improves clean accuracy substantially (65.15\% and 48.79\%) and CRI (0.787 and 0.666), but WCR remains comparable (0.796 and 0.665). 
These results suggest that extraction quality may be a bottleneck in this pipeline configuration, though a multi-OCR comparison is needed to isolate this factor.


\paragraph{Finding 5: Prompt-induced CoT improves accuracy but not worst-case robustness.}
Table~\ref{tab:cot_comparison} compares standard direct-answer prompting with prompt-induced CoT on three open-source \textit{non-thinking} models, isolating prompting effects from native thinking architectures discussed in Finding~2. 
On OCR1.0, CoT consistently improves clean accuracy by 17--30 points and raises CRI by 0.08--0.15. 
On OCR2.0, however, the gains are less robust: although clean accuracy still increases by 5--27 points, worst-case retention changes little or even declines, as in Qwen3-VL-8B where WCR drops from 0.632 to 0.594. 
Together with Finding~2, this suggests that in the evaluated models, extended reasoning mainly improves clean accuracy and does not consistently improve worst-case retention.

\subsection{Pilot Analysis on Compound Perturbations}

Real-world document images often contain more than one type of visual degradation. We additionally conduct a small scale compound perturbation experiment to examine whether single perturbation results remain informative when multiple degradations are applied together.

We use GPT-5.1 on OCR1.0, where the clean accuracy is 61.62\%. The experiment considers five perturbations at medium severity: glass blur, elastic, pixelate, color shift, and snow. Pixelate is used instead of motion blur in this pilot because it produces a clearer individual degradation signal for GPT-5.1 at this severity. We evaluate all ten pairwise combinations. To quantify interaction effects, we define the degradation ratio as
\[
\mathrm{Ratio} =
\frac{A_{\mathrm{clean}} - A_{\mathrm{compound}}}
{\sum_i \max(A_{\mathrm{clean}} - A_i, 0)},
\]
where $A_{\mathrm{clean}}$ is the clean accuracy, $A_{\mathrm{compound}}$ is the accuracy under the compound perturbation, and $A_i$ is the accuracy under each corresponding single perturbation.

The pairwise results show mixed interaction patterns. 
The largest degradation appears under pixelate + snow, where accuracy drops to 48.96\%, corresponding to a 12.66-point decrease from clean accuracy and a degradation ratio of 1.91. 
This suggests that texture damaging perturbation and occlusion perturbation may interact more strongly in this setting. 
Other combinations, such as color shift + snow, are closer to additive. 
Several pairs involving weaker perturbations, including elastic or glass blur, show sub-additive behavior. We further conduct a pilot study on triple compound perturbations, with detailed results reported in Appendix~\ref{sec:appendix_triple_perturbation}.

\section{Conclusion}

We present OCR-Robust, a benchmark for evaluating OCR reasoning robustness under controlled visual perturbations.  
Our evaluation of 18 models shows that closed-source models are generally more robust, 
while higher clean accuracy, OCR+LLM decomposition, and chain-of-thought prompting do not necessarily prevent worst-case failures under corruption.
Overall, our results suggest that OCR reasoning robustness should be treated as a distinct evaluation problem, 
requiring more robust reasoning mechanisms for future OCR-capable VLMs.

\section*{Limitations}

The dataset (812 samples) is moderate in size and language coverage. 
The perturbation selection pilot study uses a narrow basis (100 questions, 6 models, single source dataset), and alternative perturbation subsets with comparable diagnostic value are plausible. 
We evaluate 5 perturbation types individually, whereas real world degradations often co-occur; compound perturbations remain future work. Our zero-shot evaluation protocol may underestimate models' robustness under optimized prompting. 
The pipeline evaluation is limited to PaddleOCR-VL-1.5 as the sole OCR engine. 
Due to budget constraints, closed-source coverage is limited to GPT-5.x and Gemini-3-Flash. 
Finally, although all annotations undergo human review, this introduces a potential data contamination risk.

\bibliography{custom}
\appendix
\section{Details on Experimental Setup}

\subsection{Perturbation Implementation}
\label{sec:perturbation_setup}

Each of the five selected perturbations is applied at three severity levels, calibrated to produce perceptually meaningful degradation while maintaining text legibility at lower levels. All perturbations are implemented using OpenCV and PIL, with random seeds fixed per sample for reproducibility.

\paragraph{Selected Perturbations for Final Benchmark}

\textbf{Glass Blur:} Simulates the effect of viewing text through textured glass. We apply a random displacement field generated by Gaussian-filtered noise, followed by Gaussian blur. Severity levels use perturbation radius: 2, 2.5, 3 pixels.

\textbf{Color Shift:} Introduces chromatic aberration by shifting RGB channels independently. Each channel is translated by a random offset in a random direction. Severity levels use channel offset: 3, 4, 5 pixels.

\textbf{Elastic Transform:} Applies smooth geometric deformation via displacement fields generated from Gaussian-filtered random noise. Deformation strength $\alpha$ controls the magnitude of displacement. Severity levels: $\alpha = 10, 15, 20$.

\textbf{Motion Blur:} Simulates camera motion during capture by convolving with a linear motion kernel at a random angle. Severity levels use kernel size: 5, 6, 7 pixels.

\textbf{Snow:} Overlays semi-transparent snowflake patterns generated from random noise, simulating weather-induced occlusion. Severity levels use layer intensity: 0.1, 0.2, 0.3.

All perturbations preserve the original image dimensions and format. Perturbed images are saved in PNG format to avoid compression artifacts.

\paragraph{Additional Perturbations Evaluated in Pilot Study}
The following 13 perturbations were evaluated but not selected for the final benchmark:

\textbf{Noise Family:}
\textit{Gaussian Noise} adds pixel-wise Gaussian noise with standard deviation $\sigma = 15, 25, 35$.
\textit{Shot Noise} simulates photon counting noise via Poisson distribution with $\lambda = 20, 40, 60$.
\textit{Impulse Noise} randomly replaces pixels with black or white (salt-and-pepper) at rates 0.01, 0.02, 0.03.

\textbf{Blur Family:}
\textit{Defocus Blur} applies disk-shaped blur kernels with radius 2, 3, 4 pixels.
\textit{Zoom Blur} simulates radial motion blur from camera zoom with strength 1.02, 1.04, 1.06.

\textbf{Weather Family:}
\textit{Fog} overlays semi-transparent white noise layers with opacity 0.3, 0.5, 0.7.
\textit{Frost} applies crystalline patterns via Perlin noise with intensity 0.2, 0.4, 0.6.

\textbf{Digital Family:}
\textit{Brightness} adjusts luminance by factors 0.7, 0.5, 0.3 (darkening).
\textit{Contrast} reduces contrast by factors 0.7, 0.5, 0.3.
\textit{Pixelate} downsamples and upsamples with block sizes 4, 6, 8 pixels.

\textbf{Color \& Geometric Family:}
\textit{Glare} overlays elliptical gradient patterns simulating light reflection with intensity 0.4, 0.6, 0.8.
\textit{Moir\'{e}} applies sinusoidal interference patterns with frequency 0.1, 0.15, 0.2.
\textit{Slight Rotation} rotates images by 2, 4, 6 degrees with border padding.

\paragraph{LPIPS-Controlled Severity Calibration}
In the pilot study, perturbation severity was controlled using LPIPS~\citep{zhang2018lpips} (Learned Perceptual Image Patch Similarity) distance to ensure perceptually uniform degradation across perturbation types. For each perturbation, we performed binary search over parameter ranges to find settings yielding target LPIPS distances of 0.05, 0.10, 0.15, 0.20 from the clean image. This approach ensures that severity levels are comparable across different perturbation families. For the final benchmark, we use fixed parameter values (reported above) that approximate these LPIPS targets while maintaining computational efficiency.

\subsection{Prompt Templates}
\label{sec:prompt_templates}

All models receive identical prompts with four components: (1) the input image (for MLLMs) or extracted text (for pipelines), (2) the question text, (3) a format hint specifying the expected answer structure, and (4) an instruction to output only the final answer without explanation.

\begin{figure}[h]
\centering
\begin{tcolorbox}[colback=gray!10, colframe=black, rounded corners, width=0.9\columnwidth, boxrule=0.5pt]
\small
\textbf{Prompt Template for MLLMs:}

\vspace{6pt}
[Image]

\vspace{6pt}
Question: \{question\_text\}

\vspace{6pt}
\{format\_hint\}

\vspace{6pt}
Directly output the answer only, without any explanation.
\end{tcolorbox}
\caption{Evaluation prompt template for multimodal language models.}
\label{fig:prompt_mllm}
\end{figure}

\begin{figure}[h]
\centering
\begin{tcolorbox}[colback=gray!10, colframe=black, rounded corners, width=0.9\columnwidth, boxrule=0.5pt]
\small
\textbf{Example 1: Monetary Value}

\vspace{6pt}
Question: What is the total amount on this receipt?

\vspace{6pt}
The composition of the final answer should be: \$ + Integer

\vspace{6pt}
Directly output the answer only, without any explanation.
\end{tcolorbox}
\caption{Example prompt for monetary value questions.}
\label{fig:prompt_monetary}
\end{figure}

\begin{figure}[h]
\centering
\begin{tcolorbox}[colback=gray!10, colframe=black, rounded corners, width=0.9\columnwidth, boxrule=0.5pt]
\small
\textbf{Example 2: Free-form Answer}

\vspace{6pt}
Question: What is the main topic of this document?

\vspace{6pt}
Answer in English.

\vspace{6pt}
Directly output the answer only, without any explanation.
\end{tcolorbox}
\caption{Example prompt for free-form questions.}
\label{fig:prompt_freeform}
\end{figure}

\begin{figure}[h]
\centering
\begin{tcolorbox}[colback=gray!10, colframe=black, rounded corners, width=0.9\columnwidth, boxrule=0.5pt]
\small
\textbf{Example 3: Numerical Calculation}

\vspace{6pt}
Question: Based on the chart, what is the percentage increase from 2020 to 2021?

\vspace{6pt}
The composition of the final answer should be: Integer + \%

\vspace{6pt}
Directly output the answer only, without any explanation.
\end{tcolorbox}
\caption{Example prompt for numerical calculation questions.}
\label{fig:prompt_numerical}
\end{figure}

\begin{figure*}[t]
\centering
\includegraphics[width=0.48\textwidth]{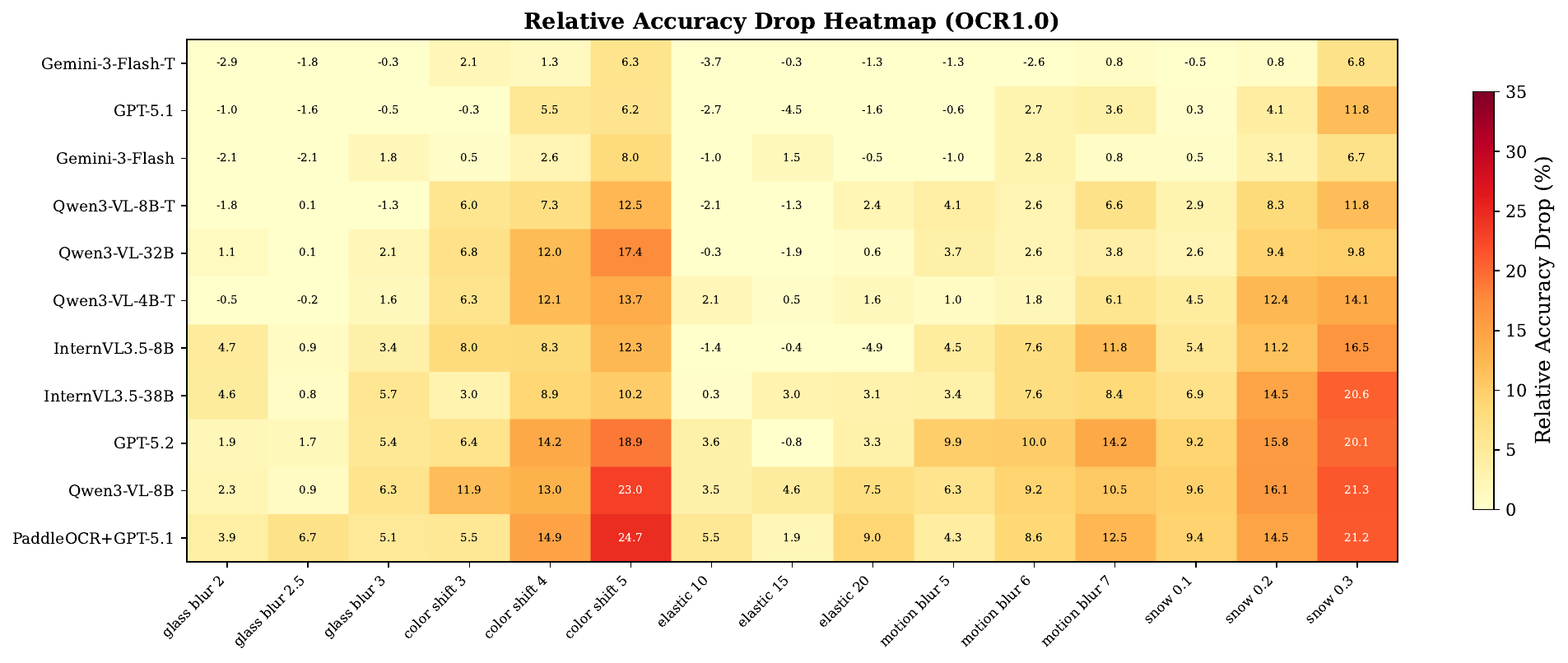}
\hfill
\includegraphics[width=0.48\textwidth]{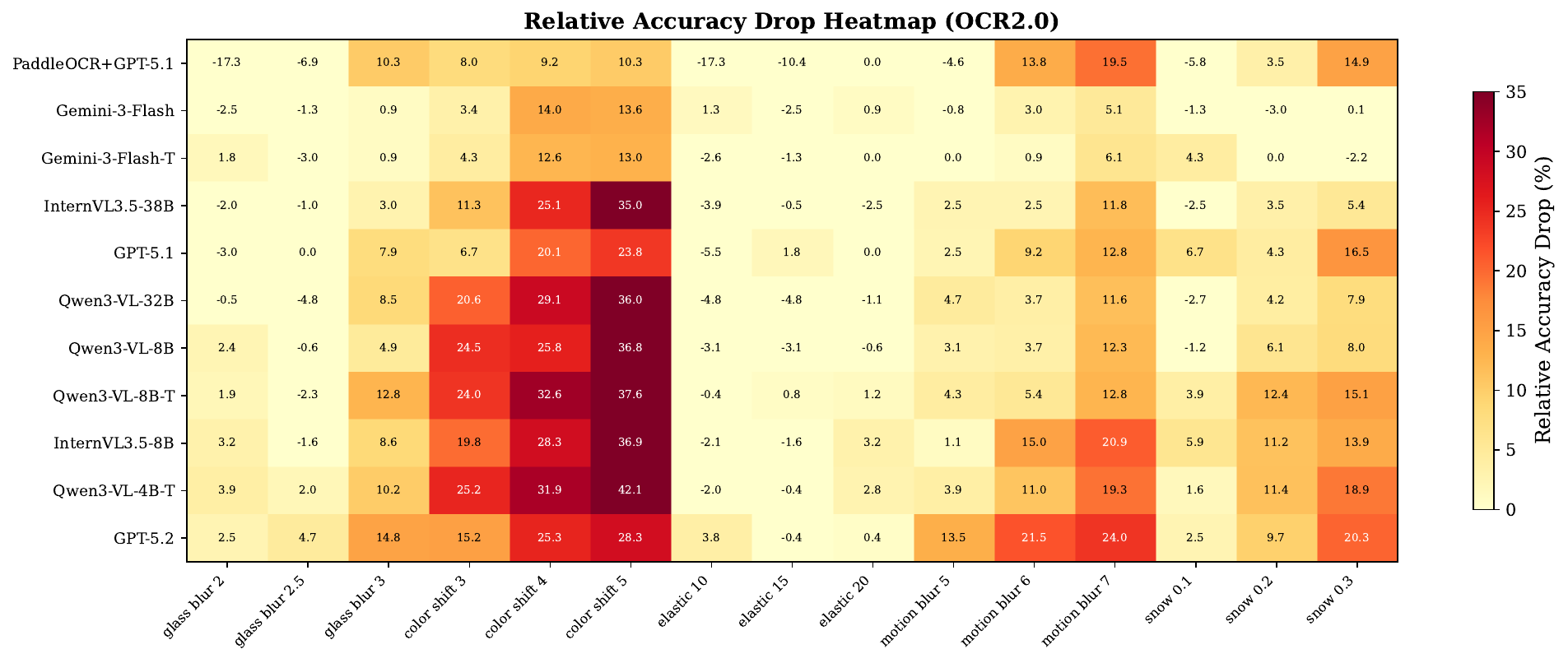}
\caption{Relative accuracy drop (\%) heatmap across all perturbation conditions. Darker cells indicate larger drops. Negative values indicate that the perturbed accuracy is higher than the clean accuracy.}
\label{fig:heatmap_drop}
\end{figure*}

\subsection{Evaluation Protocol}
\label{sec:eval_protocol}

\paragraph{Answer Normalization}
Before matching, both ground-truth and predicted answers undergo normalization: (1) convert to lowercase, (2) remove punctuation except decimal points and currency symbols, (3) strip leading/trailing whitespace, (4) collapse multiple spaces to single space. 

\paragraph{Exact Match with GPT-4o Judge Fallback}
We first attempt normalized exact match. If this fails and the question type is not multiple-choice, we invoke GPT-4o as a semantic judge. The judge is used for approximately 15--20\% of samples across the benchmark, primarily for free-form answers where phrasing varies but meaning is preserved.
To evaluate the stability of GPT-4o-based answer verification, we manually reviewed all 367 cases where exact match failed and GPT-4o was used for semantic judgment. 
The agreement between GPT-4o and human annotators reached 99.18\%. Cohen's $\kappa$ was 0.967 on OCR1.0 and 0.903 on OCR2.0. 
Only 3 out of the 367 judgments were marked as inconsistent during manual inspection.

\begin{figure}[h]
\centering
\begin{tcolorbox}[colback=gray!10, colframe=black, rounded corners, width=0.9\columnwidth, boxrule=0.5pt]
\small
\textbf{GPT-4o Judge Prompt:}

\vspace{6pt}
Question: \{question\}

Ground truth: \{gt\_answer\}

Prediction: \{pred\_answer\}

\vspace{6pt}
Are these two answers semantically equivalent? Answer only "yes" or "no".
\end{tcolorbox}
\caption{Prompt template for GPT-4o semantic judge.}
\label{fig:prompt_judge}
\end{figure}

\paragraph{Format Hints}
Each question includes a format hint derived from the answer type (e.g., ``The composition of the final answer should be: \$ + Integer'' for monetary values). These hints are appended uniformly to all models' prompts. While this may favor models trained on structured output formats, we observe no systematic bias: models with similar architectures (e.g., Qwen3-VL-4B vs. 8B) show performance gaps larger than hint-induced variance, and closed-source models maintain advantages across all answer types.

\subsection{Perturbation Selection Analysis}
\label{sec:bootstrap_stability}

To verify that our perturbation selection is not an artifact of the specific pilot sample, 
we perform a bootstrap stability analysis. We resample the 100 pilot questions with replacement 1,000 times. 
For each bootstrap sample, we recompute MRD, SEP, and MON for all 18 perturbations and re-run the greedy selection procedure. 
Table~\ref{tab:bootstrap_stability} reports the selection frequency for each perturbation.

\begin{table}[h]
\centering
\small
\begin{tabular}{lcc}
\hline
\textbf{Perturbation} & \textbf{Frequency} & \textbf{Selection Rate} \\
\hline
\textbf{Elastic}* & 984/1000 & 98.4\% \\
\textbf{Snow}* & 858/1000 & 85.8\% \\
\textbf{Glass Blur}* & 856/1000 & 85.6\% \\
\textbf{Motion Blur}* & 722/1000 & 72.2\% \\
\textbf{Color Shift}* & 442/1000 & 44.2\% \\
\hline
Defocus Blur & 287/1000 & 28.7\% \\
Zoom Blur & 284/1000 & 28.4\% \\
Impulse Noise & 222/1000 & 22.2\% \\
Shot Noise & 164/1000 & 16.4\% \\
Glare & 69/1000 & 6.9\% \\
Gaussian Noise & 47/1000 & 4.7\% \\
Slight Rotation & 22/1000 & 2.2\% \\
Moir\'{e} & 22/1000 & 2.2\% \\
Frost & 18/1000 & 1.8\% \\
Fog & 3/1000 & 0.3\% \\
\hline
\end{tabular}
\caption{Bootstrap stability analysis of perturbation selection (N=1000 resamples). * denotes the 5 perturbations selected on the full pilot sample.}
\label{tab:bootstrap_stability}
\end{table}

\begin{table*}[t]
\centering
\small
\resizebox{\textwidth}{!}{
\begin{tabular}{clcccc}
\toprule
Rank & Model & Clean Acc. [95\% CI] & RCR [95\% CI] & WCR [95\% CI] & CRI [95\% CI] \\
\midrule
1 & Gemini-3-Flash-Thinking & 78.9 [74.7, 82.5] & 0.999 [0.971, 1.027] & 0.926 [0.889, 0.964] & 0.901 [0.881, 0.918] \\
2 & Gemini-3-Flash & 80.8 [77.5, 84.3] & 0.985 [0.965, 1.006] & 0.916 [0.881, 0.950] & 0.900 [0.881, 0.916] \\
3 & Qwen3-VL-32B & 73.1 [68.9, 77.2] & 0.953 [0.924, 0.985] & 0.826 [0.774, 0.875] & 0.832 [0.810, 0.853] \\
4 & Llama-4-Scout-17B & 67.8 [63.9, 71.8] & 0.940 [0.909, 0.973] & 0.861 [0.801, 0.914] & 0.819 [0.796, 0.841] \\
5 & GPT-5.2 & 74.7 [70.6, 78.9] & 0.911 [0.880, 0.943] & 0.794 [0.745, 0.841] & 0.815 [0.790, 0.836] \\
6 & GPT-5.1 & 60.5 [56.2, 64.7] & 0.985 [0.948, 1.028] & 0.885 [0.817, 0.947] & 0.808 [0.782, 0.832] \\
7 & InternVL3.5-38B & 63.0 [58.9, 67.2] & 0.934 [0.897, 0.968] & 0.799 [0.724, 0.862] & 0.777 [0.750, 0.803] \\
8 & InternVL3.5-8B & 57.6 [53.4, 62.2] & 0.940 [0.899, 0.982] & 0.830 [0.759, 0.893] & 0.766 [0.740, 0.789] \\
9 & Qwen3-VL-8B & 54.3 [49.7, 58.9] & 0.901 [0.859, 0.945] & 0.762 [0.696, 0.821] & 0.720 [0.691, 0.748] \\
10 & Qwen3-VL-4B & 49.9 [45.5, 53.9] & 0.928 [0.884, 0.975] & 0.788 [0.718, 0.854] & 0.713 [0.684, 0.739] \\
\bottomrule
\end{tabular}
}
\caption{Bootstrap estimates on OCR1.0. Confidence intervals are computed from 1,000 sample-level bootstrap resamples.}
\label{tab:bootstrap-ocr1}
\end{table*}

\begin{table*}[t]
\centering
\small
\resizebox{\textwidth}{!}{
\begin{tabular}{clcccc}
\toprule
Rank & Model & Clean Acc. [95\% CI] & RCR [95\% CI] & WCR [95\% CI] & CRI [95\% CI] \\
\midrule
1 & Gemini-3-Flash & 71.7 [68.1, 75.4] & 0.976 [0.951, 1.000] & 0.856 [0.793, 0.914] & 0.840 [0.813, 0.863] \\
2 & Gemini-3-Flash-Thinking & 69.6 [66.0, 73.6] & 0.977 [0.950, 1.005] & 0.869 [0.807, 0.926] & 0.835 [0.807, 0.862] \\
3 & GPT-5.2 & 71.7 [67.8, 75.7] & 0.876 [0.837, 0.917] & 0.720 [0.645, 0.792] & 0.766 [0.734, 0.795] \\
4 & InternVL3.5-38B & 61.4 [57.1, 65.7] & 0.941 [0.903, 0.983] & 0.649 [0.560, 0.733] & 0.721 [0.683, 0.758] \\
5 & Llama-4-Scout-17B & 59.9 [55.6, 64.4] & 0.863 [0.820, 0.912] & 0.706 [0.616, 0.790] & 0.710 [0.672, 0.742] \\
6 & GPT-5.1 & 49.5 [45.6, 53.5] & 0.933 [0.895, 0.974] & 0.764 [0.699, 0.832] & 0.704 [0.670, 0.739] \\
7 & Qwen3-VL-32B & 57.1 [53.2, 61.1] & 0.930 [0.902, 0.961] & 0.640 [0.571, 0.715] & 0.698 [0.661, 0.734] \\
8 & InternVL3.5-8B & 56.8 [53.2, 60.8] & 0.891 [0.847, 0.937] & 0.631 [0.535, 0.717] & 0.683 [0.646, 0.720] \\
9 & Qwen3-VL-8B & 49.3 [45.9, 52.9] & 0.919 [0.883, 0.956] & 0.631 [0.560, 0.708] & 0.660 [0.623, 0.697] \\
10 & Qwen3-VL-4B & 49.5 [45.9, 53.2] & 0.843 [0.798, 0.890] & 0.557 [0.474, 0.636] & 0.613 [0.571, 0.653] \\
\bottomrule
\end{tabular}
}
\caption{Bootstrap estimates on OCR2.0. Confidence intervals are computed from 1,000 sample-level bootstrap resamples.}
\label{tab:bootstrap-ocr2}
\end{table*}

\begin{table}[t]
\centering
\small
\begin{tabular}{lc}
\toprule
Perturbation & Selection Frequency \\
\midrule
Elastic Deformation & 98.4\% \\
Snow & 85.8\% \\
Glass Blur & 85.6\% \\
Motion Blur & 72.2\% \\
Color Shift & 44.2\% \\
Defocus Blur & 28.7\% \\
\bottomrule
\end{tabular}
\caption{Bootstrap selection frequency in the perturbation pilot study.}
\label{tab:perturbation-selection-bootstrap}
\end{table}

These analyses provide a basic check on benchmark stability and perturbation-selection sensitivity. The pilot study is still limited in scale, using 100 questions and 6 models, so the selected perturbation set should be interpreted as a controlled diagnostic choice for OCR reasoning robustness rather than a comprehensive characterization of all possible real-world corruptions.

\subsection{Data Contamination Audit}
\label{sec:appendix_contaminaiton}
To assess possible data contamination, we conduct a no-image evaluation using GPT-5.4. In this setting, models receive only the question text, without the corresponding image. The goal is to estimate whether some questions can be answered without visual evidence, and to distinguish legitimately text-answerable cases from cases that may indicate memorization or leakage.

For GPT-5.4, approximately 17\% of questions can be answered correctly without the image. We further inspect these no-image correct cases and find that many of them are legitimately answerable from the textual input alone. These include pure mathematics problems where the necessary numbers are already present in the question text, which account for approximately 36\% of no-image correct cases; symbolic geometry problems with self-contained textual conditions, especially in OCR2.0, where such cases account for about 51\% of no-image successes; and a small number of culturally or semantically inferable answers, such as inferring ``Malaysia'' from the abbreviation ``RM''.

After excluding these legitimately answerable cases, we treat the remaining no-image correct answers as suspected contamination cases. These are samples that appear to require reading visual content, but are nevertheless answered correctly without access to the image. Under this criterion, suspected contamination accounts for 10\% of the full OCR1.0 dataset, which contains 482 samples, and approximately 5\% of the full OCR2.0 dataset, which contains 330 samples. OCR2.0 shows a lower suspected contamination rate, partly because many of its no-image successes come from symbolic geometry questions whose conditions are already encoded in text.

This analysis suggests that a non-trivial portion of no-image successes can be explained by the structure of the questions themselves rather than by contamination. 
At the same time, the remaining cases provide a conservative estimate of potentially contaminated samples. 
We therefore report these results as an audit of possible contamination risk rather than as definitive evidence of data leakage.

\begin{table}[t]
\centering
\small
\begin{tabular}{@{}llcccc@{}}
\toprule
\textbf{Model} & \textbf{Prompt} & \textbf{Clean} & \textbf{RCR} & \textbf{WCR} & \textbf{CRI} \\
\midrule
\multicolumn{6}{c}{\textit{OCR1.0}} \\
\midrule
\multirow{2}{*}{Qwen3-VL-8B}
& Std & 54.15 & .903 & .770 & .722 \\
& CoT & 79.25 & .965 & .874 & .875 \\
\cmidrule(l){2-6}
\multirow{2}{*}{Qwen3-VL-4B}
& Std & 49.79 & .927 & .788 & .714 \\
& CoT & 79.67 & .942 & .814 & .849 \\
\cmidrule(l){2-6}
\multirow{2}{*}{InternVL3.5-8B}
& Std & 57.26 & .941 & .835 & .766 \\
& CoT & 74.69 & .929 & .861 & .842 \\
\midrule
\multicolumn{6}{c}{\textit{OCR2.0}} \\
\midrule
\multirow{2}{*}{Qwen3-VL-8B}
& Std & 49.39 & .921 & .632 & .660 \\
& CoT & 76.06 & .893 & .594 & .739 \\
\cmidrule(l){2-6}
\multirow{2}{*}{Qwen3-VL-4B}
& Std & 49.70 & .841 & .555 & .615 \\
& CoT & 70.91 & .910 & .624 & .738 \\
\cmidrule(l){2-6}
\multirow{2}{*}{InternVL3.5-8B}
& Std & 56.67 & .892 & .631 & .683 \\
& CoT & 61.82 & .893 & .652 & .711 \\
\bottomrule
\end{tabular}
\caption{Effect of CoT prompting on robustness.}
\label{tab:cot_comparison}
\end{table}

\subsection{Few-Shot Evaluation}
\label{sec:appendix_few_shot}
We additionally conduct a preliminary few-shot evaluation. To address this point, we conducted a preliminary few-shot experiment with GPT-5.4 on OCR1.0 under 0-shot, 1-shot, 2-shot, and 3-shot settings. The results are shown in Table~\ref{tab:fewshot_results}.

\begin{table}[t]
\centering
\small
\begin{tabular}{lcccc}
\toprule
Metric & 0-shot & 1-shot & 2-shot & 3-shot \\
\midrule
RCR & 0.917 & 0.941 & 0.916 & 0.930 \\
CRI & 0.840 & 0.853 & 0.846 & 0.838 \\
\bottomrule
\end{tabular}
\caption{Preliminary few-shot results of GPT-5.4 on OCR1.0.}
\label{tab:fewshot_results}
\end{table}

These results suggest that few-shot examples provide modest gains, but the improvement is smaller than the typical gaps observed between major model families. 

\subsection{Compound Perturbation Pilot Experiment}
\label{sec:appendix_triple_perturbation}
The triple-combination results are generally close to near-additive or sub-additive behavior. Color shift + glass blur + snow and color shift + elastic + snow both obtain a degradation ratio of 0.93, while elastic + glass blur + pixelate obtains a ratio of 1.45. These results indicate that adding more perturbation types does not necessarily lead to proportionally larger degradation, possibly because some failure modes overlap once major visual information has already been disrupted.

\begin{table}[t]
\centering
\small
\begin{tabular}{lccc}
\toprule
\textbf{Combination} & \textbf{Accuracy} & \textbf{Drop} & \textbf{Ratio} \\
\midrule
Pixelate + Snow & 48.96 & 12.66 & 1.91 \\
Elastic + Snow & 55.81 & 5.81 & 1.47 \\
Glass Blur + Pixelate & 58.30 & 3.32 & 1.23 \\
Glass Blur + Snow & 57.26 & 4.36 & 1.11 \\
Color Shift + Glass Blur & 56.64 & 4.98 & 1.09 \\
Color Shift + Snow & 52.39 & 9.23 & 1.08 \\
Color Shift + Pixelate & 55.39 & 6.23 & 0.86 \\
Color Shift + Elastic & 58.09 & 3.53 & 0.77 \\
Elastic + Pixelate & 59.75 & 1.87 & 0.69 \\
Elastic + Glass Blur & 59.75 & 1.87 & -- \\
\bottomrule
\end{tabular}
\caption{Pairwise compound perturbation results for GPT-5.1 on OCR1.0 at medium severity. N/A indicates that the ratio is undefined because the summed single perturbation drop is zero.}
\label{tab:compound-pairwise}
\end{table}

\begin{table}[t]
\centering
\small
\begin{tabular}{lccc}
\toprule
Combination & Acc. & Drop & Ratio \\
\midrule
Color Shift + Glass Blur + Snow & 53.73 & 7.89 & 0.93 \\
Color Shift + Elastic + Snow & 53.73 & 7.89 & 0.93 \\
Elastic + Glass Blur + Pixelate & 57.71 & 3.91 & 1.45 \\
\bottomrule
\end{tabular}
\caption{Triple compound perturbation results for GPT-5.1 on OCR1.0 at medium severity.}
\label{tab:compound-triple}
\end{table}

Overall, this pilot experiment suggests that compound perturbations can exhibit different interaction patterns depending on the perturbation types involved. In this setting, snow appears in several of the stronger compound degradations, and pixelate + snow produces the lowest observed accuracy. At the same time, the results are based on a single model, dataset, and severity level, so they should be treated as supporting evidence for the diagnostic value of single-perturbation analysis rather than as a general conclusion about compound degradations.

\section{Details of Data Construction}
\label{sec:appendix_data_construction}

\subsection{Human Verification}
\label{sec:appendix_human}
Two graduate student annotators, both fluent in English and Chinese, participated in the human review. The data construction pipeline consisted of VLM-assisted question generation followed by human verification. 
During review, annotators checked answer correctness, answer verifiability, OCR necessity, and image--text consistency. 
To estimate inter-annotator agreement, 20\% of the samples were double-reviewed using binary pass/reject labels, yielding Cohen's $\kappa=0.87$. Each annotator reviewed approximately 160--170 VLM-assisted generated samples, and ambiguous cases were resolved through discussion.

\subsection{Quality Control}
\label{sec:appendix_quality_control}
We apply several quality control steps during data construction.

\paragraph{Automated validation.}
All generated samples are first filtered through automated checks, including image integrity verification, format validation, and answer consistency checks.

\paragraph{Deduplication}
We perform image-level deduplication using perceptual hashing (pHash). Samples with pHash distance smaller than 5 are considered near-duplicates and removed.

\paragraph{Image quality filtering.}
We filter images using a minimum resolution threshold of $224 \times 224$ and an RMS contrast threshold larger than 15.

\subsection{Synthetic Mathematics Subset}

The synthetic mathematics subset follows a construction process similar to LogicOCR. Source problems are collected from GSM8K and TheoremQA and converted into free-form visual question-answering examples. GPT-4o is used to generate contextual visual descriptions, and GPT-4o-image renders images containing the generated context, question, and answer. Ground-truth answers are preserved from the original verified datasets.

To ensure textual fidelity after rendering, we apply PaddleOCR-based character-level recall filtering against the intended source text. Samples with recall below 0.9 are regenerated or discarded. This filtering step is used to reduce rendering artifacts and maintain consistency between the rendered image and the source problem content.

\subsection{Visual Diversity Sampling and Quality Control}

To improve visual coverage and reduce redundancy, we apply a clustering-based sampling strategy, mainly for samples derived from ChartQA~\citep{masry2022chartqa}. This procedure serves as a diversity-oriented heuristic rather than a formal guarantee of dataset diversity.

For each candidate image, we extract a 2048-dimensional feature using an ImageNet-pretrained ResNet, taking the pooled representation before the final fully connected layer. Images are preprocessed with Resize(256), CenterCrop(224), and standard normalization, and the resulting features are $\ell_2$-normalized. 
We then perform agglomerative clustering with Ward linkage and set $K=150$. Samples closest to the cluster centroids are selected to improve visual coverage and reduce near-duplicate patterns.

\subsection{Visual Examples and Visual-Domain Terminology}

\begin{figure*}[t]
\centering
\includegraphics[page=1,width=0.7\textwidth]{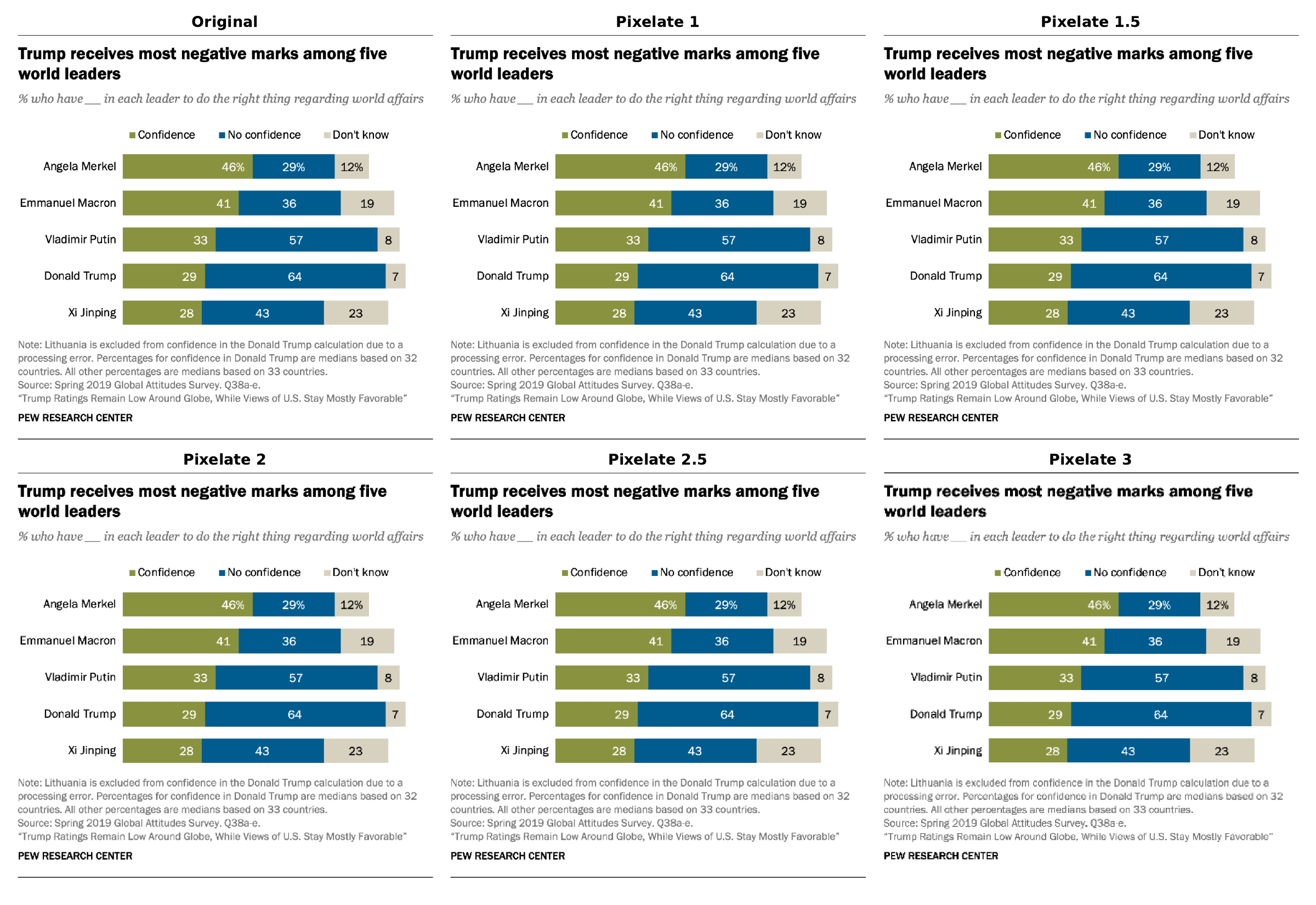}

\vspace{0.5em}

\includegraphics[page=2,width=0.7\textwidth]{chapter/figures/case.pdf}

\vspace{0.5em}

\includegraphics[page=3,width=0.7\textwidth]{chapter/figures/case.pdf}
\caption{Representative images from OCR2.0.}
\label{fig:cases}
\end{figure*}

To improve interpretability for readers from the ACL/NLP community, 
we provide representative examples from OCR-Robust as Figure~\ref{fig:cases}, including the input image, question, and ground-truth answer. 
We also include a perturbation visualization figure showing the selected perturbation types across different severity levels.

We also provide brief explanations of visual-domain terms. 
For example, \emph{channel offset} refers to shifting RGB color channels relative to each other, which resembles color misalignment in printing or low-quality cameras. 
\emph{Kernel size} denotes the size of the local pixel window used by a blur filter, 
where a larger kernel generally produces stronger blur. 

\section{Additional Figures}

\begin{figure*}[h]
\centering
\includegraphics[width=\textwidth]{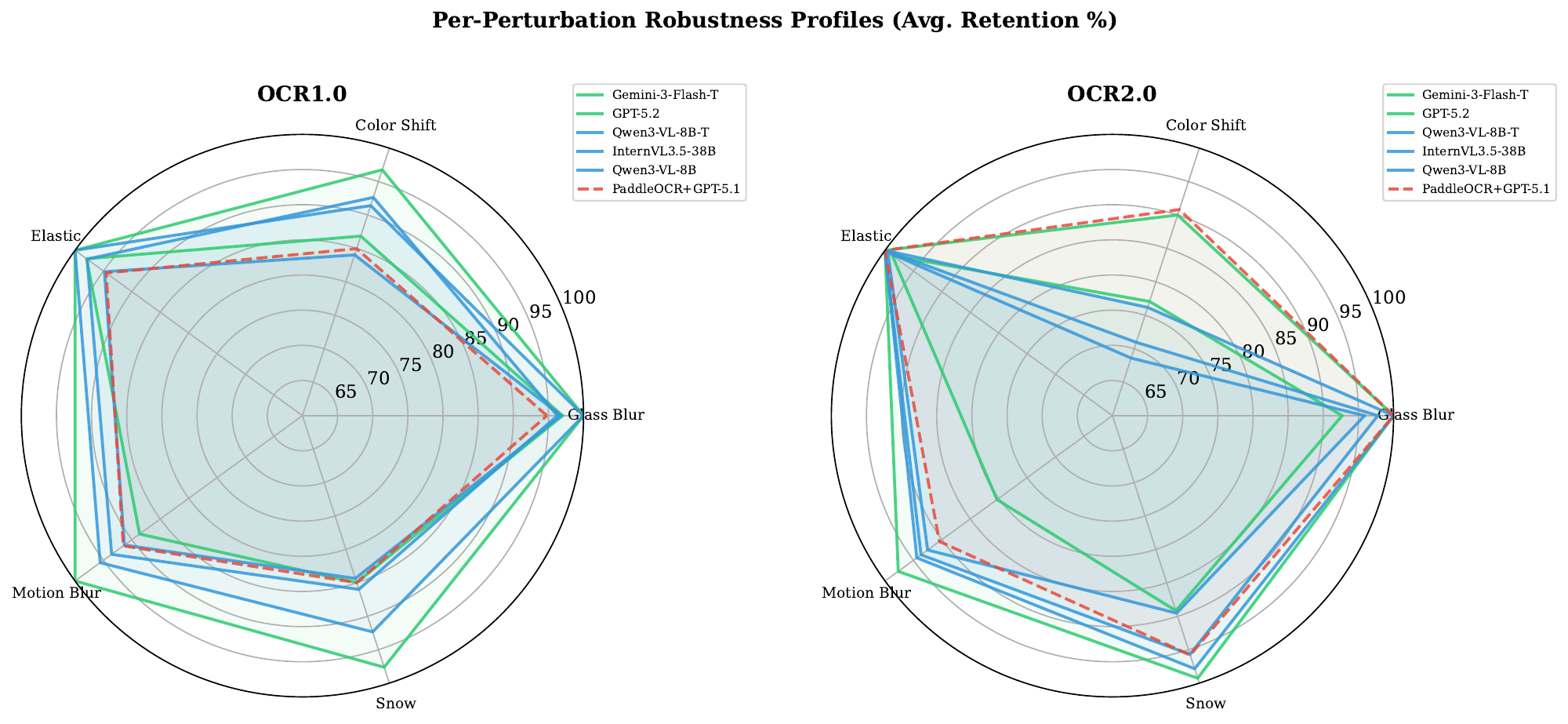}
\caption{Per-perturbation robustness profiles showing average retention (\%) across five perturbation types. Closed-source models (green) maintain near-circular profiles, while pipelines (red dashed) exhibit pronounced weaknesses.}
\label{fig:radar_profiles}
\end{figure*}


\begin{figure*}[h]
\centering
\includegraphics[width=\textwidth]{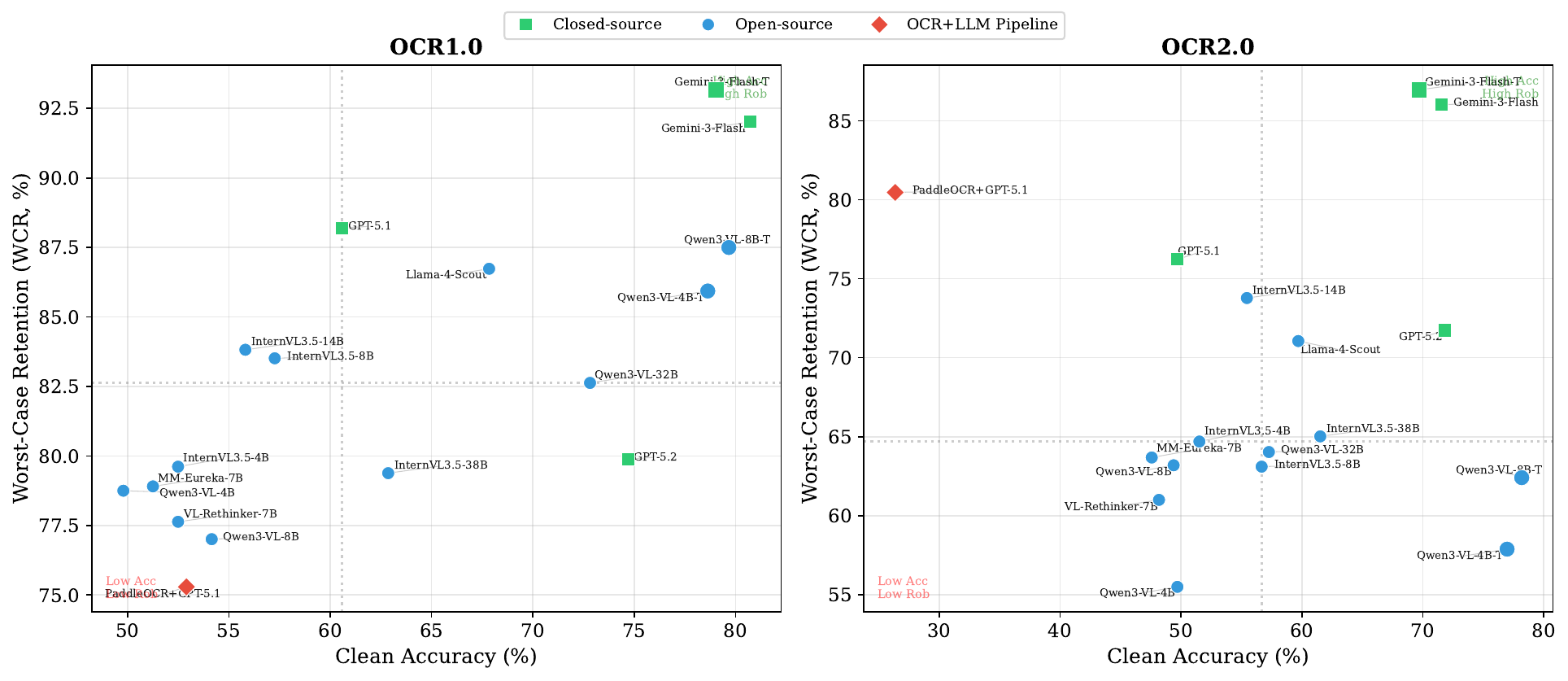}
\caption{Clean accuracy vs.\ worst-case retention (WCR). Models in the upper-right quadrant achieve both high accuracy and strong robustness. Dashed lines indicate medians.}
\label{fig:clean_vs_wcr}
\end{figure*}

\end{document}